\documentclass[a4paper]{ociamthesis}
\usepackage{graphicx}
\usepackage{colortbl}
\usepackage{enumitem}
\usepackage{amsthm}
\usepackage{yhmath}
\usepackage[square]{natbib}
\usepackage[most]{tcolorbox}
\usepackage{dsfont}
\definecolor{bordeau}{rgb}{0.3515625,0,0.234375}
\definecolor{prune}{RGB}{99,0,60}
\definecolor{gris1}{RGB}{49,62,72}
\definecolor{gris2}{RGB}{124,135,143}
\definecolor{gris3}{RGB}{213,218,223}
\definecolor{rouge}{RGB}{198,11,70}
\definecolor{rose}{RGB}{237,20,91}
\definecolor{orange1}{RGB}{238,52,35}
\definecolor{orange2}{RGB}{243,115,32}
\definecolor{violet1}{RGB}{124,42,144}
\definecolor{violet2}{RGB}{125,106,175}
\definecolor{marron}{RGB}{198,103,29}
\definecolor{jaune}{RGB}{254,188,24}
\definecolor{bleu1}{RGB}{0,78,125}
\definecolor{bleu2}{RGB}{14,135,201}
\definecolor{bleu3}{RGB}{0,148,181}
\definecolor{bleu4}{RGB}{70,195,210}
\definecolor{vert1}{RGB}{0,128,122}
\definecolor{vert2}{RGB}{64,183,105}
\definecolor{vert2}{RGB}{64,183,105}
\definecolor{vert3}{RGB}{140,198,62}
\definecolor{vert4}{RGB}{213,223,61}

\tcbset {
  base/.style={
    arc=0mm, 
    bottomtitle=0.5mm,
    boxrule=0.1mm,
    coltitle=black, 
    left=2.5mm,
    leftrule=1mm,
    right=3.5mm,
    title={#1},
    toptitle=0.75mm,
    parbox=false,
    fonttitle=\bfseries,
  }
}

\newtcolorbox{definitionbox}[1]{
  breakable, base={#1}, 
  colback=vert1!10, 
  colframe=vert1, 
  colbacktitle=vert1!50, 
}

\newtcolorbox{examplebox}[1]{
  breakable, base={#1}, 
  colback=gris3!25, 
  colframe=gris1, 
  colbacktitle=gris1!50, 
}

\newtcolorbox{resultbox}[1]{
  breakable, base={#1}, 
  colback=prune!10, 
  colframe=prune, 
  colbacktitle=prune!50, 
}

\usepackage{titlesec}

\usepackage{setspace}

\begin{document}
\setlength{\textbaselineskip}{15pt plus2pt}

\titlespacing{\section}{0pt}{1cm}{0.2cm}
\titlespacing{\subsection}{0pt}{0.5cm}{0.1cm}
\titlespacing{\subsubsection}{0pt}{0.5cm}{0.1cm}
\setlength{\frontmatterbaselineskip}{17pt plus1pt minus1pt}

\setlength{\baselineskip}{\textbaselineskip}

\counterwithout{section}{chapter}

\flushbottom

\setcounter{page}{1}

\textbf{\chapter*{\centering\color{black}Causal Inference Tools for a Better Evaluation of Machine Learning\\{\Large\textmd{Michaël Soumm}}}}

\thispagestyle{fancy}

\section{Introduction and Motivation}
This paper presents a comprehensive framework for applying rigorous statistical methods to the evaluation of machine learning models. While machine learning models often focus on predictive accuracy, understanding the causal relationships and factors influencing model performance is fundamental for advancing the field. The techniques presented here allow us to move beyond simple performance metrics and understand the underlying mechanisms of our models.
The methods discussed in this work form the foundation of statistical causal modeling, providing tools to rigorously analyze and interpret ML results. This approach draws primarily from established econometric techniques, which offer a robust framework for understanding complex relationships in data. The content is largely based on seminal works in the field of econometrics, including "Mostly Harmless Econometrics" by \citet{angrist2009mostly}, "Econométrie: méthodes et applications" by \citet{crepon2005}, and "Econometrics" by \citet{wooldridge2013introductory}.

By applying these causal inference methods to machine learning problems, we aim to improve our models and gain deeper insights into the factors driving their performance across different datasets and scenarios.
\subsection{The Need for Rigorous Statistical Analysis in Machine Learning}

The field of machine learning has seen rapid advancements in recent years, with new models and techniques frequently outperforming existing methods. However, the complexity of these models and the multitude of factors influencing their performance necessitate a more rigorous approach to analysis and evaluation.

Traditional machine learning metrics, such as accuracy, precision, recall, or F1 score, while valuable, often provide a limited view of a model's capabilities. They may obscure the underlying factors contributing to performance improvements or fail to account for statistical uncertainties. This limitation becomes particularly problematic when comparing models across different datasets, architectures, or training regimes.

Rigorous statistical analysis offers a framework to address these challenges. By employing techniques from econometrics and statistical inference, we can:
\begin{itemize}[noitemsep]
    \item Isolate the effects of individual factors on model performance
    \item Quantify the uncertainty in our performance estimates
    \item Control for confounding variables that might skew results
    \item Test hypotheses about the relative importance of different model components or training strategies
    \item Provide more robust and generalizable insights into model behavior
\end{itemize}

Furthermore, statistical analysis allows us to move beyond mere performance comparisons and toward a better understanding of the causal mechanisms underlying our models. This approach advances the theoretical foundations of machine learning evaluation, which in turn provide clues for future research directions.

\begin{examplebox}{Example: Beating the State-of-the-Art}
Consider a scenario where a new Deep Learning recognition model achieves a 5\% increase in accuracy over the state-of-the-art. This improvement could be attributed to multiple factors:
\begin{enumerate}
    \item A new neural network architecture
    \item An increased number of parameters
    \item A larger or more diverse training dataset
    \item New data augmentation techniques
    \item A different optimization algorithm
\end{enumerate}

Without rigorous analysis, determining the true source of improvement is challenging. Statistical methods can help isolate the effects of each factor. This allows for statements such as: \textit{"Controlling for model size and dataset characteristics, our novel architecture contributes to a 2.3\% increase in accuracy"}
\end{examplebox}

By adopting these rigorous statistical approaches, we can enhance the reliability and interpretability of machine learning research. This is particularly important in high-stakes applications where understanding model behavior and reliability is necessary. Moreover, it facilitates more meaningful comparisons between different approaches and helps identify the most promising avenues for further investigation.

In the subsequent sections, we will describe the specific statistical techniques that enable this level of rigorous analysis, providing the theoretical foundation and practical tools that can be used throughout applications.

\subsection{Limitations of Simple Performance Metrics}
While simple performance metrics are widely used in machine learning evaluation, they often fall short in providing a comprehensive understanding of model behavior and performance. These limitations become particularly apparent in complex machine learning systems and when dealing with real-world applications. Here, we discuss several limitations:
\begin{enumerate}[noitemsep]
    \item \textbf{Lack of statistical rigor:} Simple metrics typically report point estimates without confidence intervals or measures of statistical significance. This can lead to over-interpretation of small differences between models, which may not be statistically meaningful.
    \item \textbf{Insensitivity to data distribution:} Metrics like accuracy can be misleading when dealing with imbalanced datasets. A model achieving 99\% accuracy on a dataset with a 99:1 class imbalance might simply be predicting the majority class for all instances \citep{ClassImabalanceImpact}.
    \item \textbf{Failure to capture model uncertainty:} Traditional metrics do not account for the model's confidence in its predictions. This is particularly important in applications where knowing when a model is uncertain is as important as its accuracy \citep{gal2016uncertainty}.
    \item \textbf{Inability to explain performance differences:} When comparing models, simple metrics can tell us which model performs better, but not why. They fail to provide insights into the factors contributing to performance differences \citep{lipton2018mythos}.
    \item \textbf{Lack of causal understanding:} Performance metrics do not reveal causal relationships between model components, training strategies, or data characteristics and the resulting performance \citep{pearl2018book}.
    \item \textbf{Vulnerability to dataset bias:} Models can achieve high performance on test sets while still failing to generalize to real-world scenarios due to hidden biases in the evaluation data \citep{torralba2011unbiased}.
\end{enumerate}

To illustrate these limitations, consider the following example:

\begin{examplebox}{Example: Limitations of Accuracy in Face Recognition}
A face recognition system achieves 95\% accuracy on a test set. However, this metric alone fails to reveal that:

\begin{itemize}
    \item The system's performance varies significantly across different demographic groups.
    \item Most errors occur in low-light conditions, an important factor for real-world deployment.
    \item The system is overly confident in its incorrect predictions, potentially leading to errors in high-stakes applications.
    \item The test set lacks diversity in age groups, potentially overestimating the model's generalization ability.
\end{itemize}

The simple accuracy metric does not capture these insights, essential for understanding the model's true capabilities and limitations.
\end{examplebox}

To address these limitations, researchers increasingly turn to more sophisticated evaluation techniques. These include statistical hypothesis testing, confidence interval estimation, analysis of variance (ANOVA), and causal inference methods. By employing these advanced techniques, we can gain a more nuanced and reliable understanding of model performance, guiding both the development of better models and their responsible deployment in real-world scenarios.

\subsection{Econometric Regressions as Tools for Understanding Relationships}
Econometric regression techniques offer powerful tools for analyzing and interpreting complex relationships in machine learning systems. These methods, originally developed in economics and social sciences, provide a rigorous framework for quantifying the impact of various factors on model performance and behavior. In the context of machine learning, econometric approaches can help researchers move beyond simple performance comparisons to gain deeper insights into the underlying mechanisms driving model outcomes.
Regression analysis allows us to model the relationship between a dependent variable (such as model performance) and one or more independent variables (such as model architecture, dataset characteristics, or training hyperparameters). By doing so, we can:
\begin{enumerate}
\item \textbf{Isolate individual effects:} Regression models can control for multiple factors simultaneously, allowing us to estimate the impact of a single variable while holding others constant. This is crucial in machine learning, where performance improvements may result from the interplay of numerous factors.
\item \textbf{Quantify uncertainty:} Econometric methods provide measures of statistical significance and confidence intervals, enabling researchers to assess the reliability of their findings and avoid overinterpreting noise in the data.

\item \textbf{Test hypotheses:} Regression analysis offers a formal framework for testing hypotheses about the relationships between variables, allowing researchers to validate or refute theoretical predictions about model behavior.

\item \textbf{Identify non-linear relationships:} Advanced regression techniques can capture complex, non-linear relationships between variables, providing insights into subtle interactions within machine learning systems.

\item \textbf{Handle heterogeneity:} Econometric methods can account for variations across different subgroups or contexts, revealing how model performance might differ across diverse datasets or application scenarios.
\end{enumerate}
By employing econometric regression techniques, machine learning researchers can move beyond black-box comparisons and gain a deeper understanding of their models. This approach not only enhances the interpretability of results but also provides a solid foundation for making informed decisions about model design, data collection, and optimization strategies.
Moreover, the use of econometric methods in machine learning aligns with the growing emphasis on explainable AI and responsible development of AI systems. By providing a rigorous framework for analyzing model behavior, these techniques contribute to the broader goal of creating more transparent, reliable, and trustworthy machine learning systems \citep{Balancing}.
The application of econometric methods to machine learning challenges represents a promising interdisciplinary approach. It combines the statistical rigor of econometrics with the computational power and flexibility of machine learning, potentially leading to new insights and methodologies that can benefit both fields. As machine learning systems become increasingly complex and are deployed in critical real-world applications, the need for such rigorous analytical tools becomes ever more important.

\section{Foundations of Econometrics: Ordinary Least Squares (OLS)}
Ordinary Least Squares (OLS) is a fundamental method in econometrics and serves as the foundation for many advanced statistical techniques. It provides a straightforward yet powerful approach to modeling relationships between variables, and its understanding is fundamental to being able to draw meaningful conclusions from analytic experiments.
\subsection{The Linear Regression Model}
The linear regression model is a statistical method used to model the relationship between a dependent variable and one or more independent variables.
\begin{definitionbox}{Definition: Endogenous and Exogenous Variables}
In econometrics, variables are classified as:
\begin{itemize}
\item \textbf{Endogenous variable}: The dependent variable ($Y$) that the model aims to explain or predict. It is determined within the model and is influenced by the exogenous variables.
\item \textbf{Exogenous variables}: The independent variables ($X$) that are used to explain or predict the endogenous variable. They are determined outside the model and are assumed to influence the endogenous variable.
\end{itemize}
\end{definitionbox}
Let $Y$ and $X$ be random variables modeling an endogenous and exogenous variable. Then, the simple linear model is:
\begin{equation}
   Y = \beta_0 + \beta_1 X + \varepsilon 
\end{equation}
Where:
\begin{itemize}[noitemsep]
\item $\beta_0 \in \mathbb{R}$ is the called \textbf{\textit{intercept term}}
\item $\beta_1 \in \mathbb{R} $ is the \textit{\textbf{slope coefficient}}
\item $\varepsilon$ is the \textbf{\textit{error}} random variable
\end{itemize}
In practice, however, we often want to consider multiple explanatory factors, leading to the full linear model:
\begin{definitionbox}{Definition: The Linear Regression Model}
Let $Y$ be an endogenous variable and $X_1, ..., X_K$ be exogenous variable. The full Linear Regression Model is
\begin{equation}
\label{eq:linear}
    Y = \beta_0 + \beta_1 X_1 + \beta_2 X_2 + ... + \beta_K X_K + \varepsilon
\end{equation}
where $\beta_1, \beta_2, ..., \beta_k$ are their regression coefficients and $\varepsilon$ is models the error
\end{definitionbox}
The model in equation $\ref{eq:linear}$ should be unfamiliar to anyone with a Machine Learning background, since linear regression is the building block of many ML algorithms, including neural networks. However, whereas in applied machine learning, using this model as a predictive tool is common, in econometrics, the focus is put on finding the right assumptions that allow us to interpret the coefficients $\beta_k$ as causal effects and draw explanatory conclusions.
\subsection{Interpreting the Coefficients}
Several key assumptions must hold for the $\beta_k$ coefficients in the linear regression model to be interpreted as causal effects. These assumptions are fundamental to the Gauss-Markov theorem and are often referred to as the Classical Linear Regression Model (CLRM) assumptions.
\begin{resultbox}{Classical Linear Regression Model (CLRM) assumptions}
\begin{itemize}
    \item \textbf{Correct Specification}: the regression equation contains all of the relevant predictors, including any necessary transformations. That is, the model has no missing, redundant, or extraneous predictors.
    \item \textbf{Linearity in Parameters}: The relationship between $Y$ and $X_k$ is linear in the parameters $\beta_k$. This doesn't mean $X_k$ and $Y$ must have a linear relationship, but rather that the parameters enter the equation linearly.
    \item \textbf{Exogeneity}: $\mathbb{E}[\varepsilon|X_k] = 0]$ for all $k$. In other words, there are no omitted variables that are correlated with both the dependent variable and the independent variables. This is needed for causal interpretation as it ensures that unobserved factors do not bias the estimated effects.
    \item \textbf{Normality of the Error Term}: $\varepsilon \sim \mathcal{N}(0, \sigma^2)$. While not strictly necessary for unbiasedness, the assumption that the error term is normally distributed allows for valid inference (hypothesis testing and confidence intervals).
    \item\textbf{Homoscedasticity}: $\mathbb{V}(\varepsilon|X_k) = \sigma^2$. This assumption states that the variance of the error term is constant across all levels of the independent variables. Homoscedasticity ensures that the precision of the $\beta_k$ estimates is consistent across the range of $X_k$ values.
    \item \textbf{No Perfect Multicollinearity}: There should be no exact linear relationships among the independent variables. This ensures that we can uniquely estimate the effect of each variable.
\end{itemize}
\end{resultbox}
When all the criteria are met, we can interpret the coefficients as causal effects:
\begin{resultbox}{Linear Regression Model Interpretation}
When all CLRM assumptions are met, then each coefficient $\beta_k$ represents the \textbf{ marginal effect} of the corresponding exogenous variable $X_k$ on the endogenous variable $Y$, holding all other variables constant. Mathematically, this can be expressed as:
\begin{equation}
   \beta_k = \frac{\partial Y}{\partial X_k}
\end{equation}
\end{resultbox}
The "holding all other variables constant" condition is known as the \textit{ceteris paribus} assumption. It's essential for isolating the effect of a single variable and is a key concept in causal interpretation \citep{angrist2009mostly}.

Regression over many variables $X_k$ simultaneously allows us to disentangle the effect of each variable and, under the right assumptions, draw meaningful conclusions on the relationship between $X_k$ and $Y$. In particular, under the exogeneity and correct specification assumptions, the $\beta_k$ coefficient can be interpreted as the \textbf{causal effect} of $X_k$ on $Y$.
\subsection{Fitting the Model}
From now, let us denote by $X$ the row vector $(1, X_1, ..., X_K)$, and $\beta$ the vector of all $\beta_k$. Then, under the CLRM assumptions, we have a closed formula for the coefficients:
\begin{resultbox}{Formula for the Coefficients}
Under the exogeneity and no perfect colinearity assumptions, the expression
\begin{equation}
    \beta = \mathbb{E}\left[X^TX\right]^{-1}\mathbb{E}\left[X^TY\right]
\end{equation}
is a solution to the equation \ref{eq:linear}.
\end{resultbox}
\begin{proof}
    We have :
    \begin{align*}
        Y &=X\beta + \varepsilon\\
        X^TY &= X^TX\beta + X^T\varepsilon\\
        \mathbb{E}[X^TY] &= \mathbb{E}[X^TX]\beta + \mathbb{E}[X^T\varepsilon]\\
         \mathbb{E}[X^TY] &= \mathbb{E}[X^TX]\beta \quad\text{by exogeneity}\\
         \beta &=\mathbb{E}[X^TX]^{-1} \mathbb{E}[X^TY]
    \end{align*}
    The inversion of $\mathbb{E}[X^TX]^{-1}$ is possible under the no perfect colinearity assumption.
\end{proof}
Of course, in practice, we only have access to realizations of $(X, Y)$. We thus define the empirical counterparts of our quantities, by considering a set of $n$ realizations $(x_i, y_i) \sim (X, Y)$, and the equation:
\begin{equation}
    y_i = x_i\beta +\epsilon_i
\end{equation}
where $\epsilon_i$ is called the \textit{\textbf{residual}} the model for the observation $i$. Under the hypothesis that the $(x_i, y_i)$ are i.i.d., we should have $\epsilon_i \overset{i.i.d.}{\sim}\varepsilon$, in other terms, that the empirical residuals should be realizations of the noise variable. We can then define our estimator:
\begin{definitionbox}{The OLS Estimator}
Under the CLRM assumptions
\begin{equation}
    \widehat{\beta}^{\text{OLS}} := \left(\frac{1}{n}\sum_{i=1}^n x_i^Tx_i\right)^{-1}\left(\frac{1}{n}\sum_{i=1}^n x_i^Ty_i\right)
\end{equation}
is an estimator of $\beta$, and is called the \textbf{Ordinary Least Squares Estimator (OLS)}.
\end{definitionbox}
\begin{proof}
    Direct by the continuous mapping theorem.
\end{proof}
Through convex optimization, it it easy to show that $X^T\beta$ is the best linear approximation of $\mathbb{E}[Y|X]$ with respect to the square norm $L^2$, i.e. $\beta = \underset{b}{\arg\min}\mathbb{E}[(\mathbb{E}[Y|X]-Xb)^2]$, which justifies the name of the OLS estimator, which minimizes the empirical mean square error.

Let us denote by $\boldsymbol{X}$ the matrix of the observations $x_i$, and $\boldsymbol{y}$ the vector of observations of $y_i$. Then, the OLS estimator can be easily rewritten as 
\begin{equation}
     \widehat{\beta}^{\text{OLS}} = (\boldsymbol{X}^T\boldsymbol{X})^{-1}\boldsymbol{X}^T\boldsymbol{y}
\end{equation}
and can be viewed as the pseudo-inverse of $\boldsymbol{X}$ acting on $\boldsymbol{y}$. Plugging back $\widehat{\beta}^{\text{OLS}}$ in the equation \ref{eq:linear}, we get the predicted values vector $\hat{\boldsymbol{y}}$ defined by :
\begin{equation}
    \hat{\boldsymbol{y}} :=  \boldsymbol{X}(\boldsymbol{X}^T\boldsymbol{X})^{-1}\boldsymbol{X}^T\boldsymbol{y} \overset{not}{=}\hat{\boldsymbol{H}}\boldsymbol{y}
\end{equation}
where $\hat{\boldsymbol{H}}$ is the called the \textbf{hat matrix}, and is the projection matrix on the linear space generated by the rows of $\boldsymbol{X}$.
\subsection{Statistical Significance and Hypothesis Testing}
\subsubsection*{Individual Variables}
The OLS estimator thereby denoted simply $\hat{\beta}$, estimates the theoretical $\beta$ with a certain precision since it is a function of the realizations $(x_i, y_i)$. We can know its precision, both in a a finite sample setup and in an asymptotic setup:
\begin{resultbox}{Properties of the OLS estimator}
(Finite sample) Under the CLRM assumptions and i.i.d sampling, 
\begin{equation}
    \hat{\beta}\sim \mathcal{N}\left(\beta, \sigma^2 \left(\boldsymbol{X}^T\boldsymbol{X}\right)^{-1}\right)
\end{equation}
(Asymptotic distribution) Under the CLRM assumptions, i.i.d. sampling, and CLT moment conditions, $\hat{\beta}$ is asymptotically normal: 
\begin{equation}
    \sqrt{n}(\hat{\beta}-\beta) \overset{\mathcal{L}}{\longrightarrow} \mathcal{N}(0, \sigma^2\mathbb{E}[X^TX])
\end{equation}
\end{resultbox}
\begin{proof}
(Finite sample): We can write $\hat{\beta}$ as :
\begin{align*}
    \hat{\beta} =  \left(\frac{1}{n}\sum_{i=1}^n x_i^Tx_i\right)^{-1}\left(\frac{1}{n}\sum_{i=1}^n x_i^T (x_i\beta +\epsilon_i)\right)
\end{align*}
which simplifies to 
\begin{align*}
    \hat{\beta} = \beta + \left(\sum_{i=1}^n x_i^Tx_i\right)^{-1}\left(\sum_{i=1}^n x_i^T\epsilon_i\right)
\end{align*}
Since $\epsilon_i\overset{i.i.d.}{\sim} \varepsilon\sim \mathcal{N}(0, \sigma^2)$, $\hat{\beta}$ is normal, with tractable expected value and covariance matrix.\\
    (Asymptotic distribution): Direct by applying the CLT and calculating the variance of $\hat{\beta}$. Homoscedascity reduces the expression since $\mathbb{E}[\varepsilon^2 X^TX] =\sigma^2\mathbb{E}[X^TX]$.
\end{proof}
This formulation of the variance supposes homoscedasticity of the errors, but other formulations exist for cases where this assumption is not valid.

In practice, we have to estimate $\sigma^2\mathbb{E}[X^TX]$ from the data, leading to the empirical counterpart $s^2\hat{\mathbb{E}}[X^TX]$, where $s^2$ is the unbiased variance estimator. However, the estimation of the variance with an empirical counterpart changes the law which describes our error from a Normal distribution to a Student distribution:
\begin{definitionbox}{Definition: $t$-statistic}
Let $s^2 = \frac{\boldsymbol{\epsilon}^T\boldsymbol{\epsilon}}{n-K}$ be the unbiased estimator of $\sigma^2$, where $\boldsymbol{\epsilon}$ is the vector of the residuals. Let $se(\hat{\beta}_k) = \sqrt{s^2(\boldsymbol{X}^T\boldsymbol{X})^{-1}_{kk}}$. Then:
\begin{equation}
    t_k := \frac{\hat{\beta}_k-\beta}{se(\hat{\beta}_k)} \sim \mathcal{T}(n-K)
\end{equation}
where $\mathcal{T}(n-K)$ is the Student distribution with $n-K$ degrees of freedom. The quantity $t_k$ is called the $t$-statistic of the coefficient $\beta_k$.
\end{definitionbox}
\begin{proof}
    $s^2$ is the classical unbiased variance estimator. In particular, since the $\epsilon_i$ follow normal distributions, then we can show that $\frac{n-K}{\sigma^2}s^2 \sim \chi^2_{n-K}$ via Cochran's theorem. Rewriting the expression:
    \begin{align*}
          t_k = \frac{\hat{\beta}_k-\beta}{\sqrt{\sigma^2(\boldsymbol{X}^T\boldsymbol{X})^{-1}_{kk}}}\cdot\sqrt{\frac{\sigma^2}{s^2}}
    \end{align*}
    The first factor follows a Gaussian distribution, and the second factor can be rewritten as $\frac{1}{\sqrt{U/(n-K)}}$ where $U$ follows a $\chi^2$ distribution with $n-K$ degrees of freedom. By definition, $t_k$ follows a Student distribution with $n-K$ degrees of freedom.
\end{proof}
The $t$-statstic is the object that allows us to get confidence intervals for $\beta_k$, by looking at the likelihood of $t_k$ with respect to its theoretical Student distribution In particular, this is useful to verify if $\beta_k\neq0$, i.e. if there is a significant effect of $X_k$ on $Y$, and if yes, its sign and strength.
\begin{resultbox}{Confidence Intervals}
Let $t_{n-K,1-\alpha/2}$ be the $1-\alpha/2$-th quantile for the Student distribution with $n-K$ degrees of freedom. Then, with probability $1-\alpha$, 
\begin{equation}
    \beta_k \in \left[ \hat{\beta}_k \pm t_{n-K,1-\alpha/2}\times se(\hat{\beta}_k)\right]
\end{equation}
\end{resultbox}
A common threshold for $\alpha$ is 5\%.

If 0 is contained inside the confidence interval, it means we do not have enough data points to draw conclusions. This does not mean that $X_k$ has \textit{no} effect on $Y$, but that the noise in the observed data is too strong to even say if the effect is positive or negative. On the other hand, we can have a very narrow confidence interval, but around a small $\hat{\beta}$ value. This means that the effect is significant but small. Therefore, assessing both the strength and significance of the effects $\beta_k$ is important.

\subsubsection*{Global Significance}
The main measure of the overall fit of a linear regression model is the coefficient of determination, commonly known as $R^2$. This statistic quantifies the proportion of variance in the dependent variable that is predictable from the independent variable(s). $R^2$ ranges from 0 to 1, where 0 indicates that the model explains none of the variability of the data around its mean, and 1 indicates perfect prediction. Mathematically, $R^2$ can be defined in terms of variance and covariance:
\begin{definitionbox}{Definition: $R^2$}
    \begin{equation}
\label{eq:r2}
R^2 = \frac{\widehat{Cov}(y, \hat{y})^2}{\widehat{Var}(y)\widehat{Var}(\hat{y})} = \frac{\widehat{Var}(\hat{y})}{\widehat{Var}(y)} = 1 - \frac{\widehat{Var}(\epsilon)}{\widehat{Var}(y)}
\end{equation}
\end{definitionbox}

Formulation \ref{eq:r2} highlights that $R^2$ represents the squared correlation between the observed and predicted values, or equivalently, the ratio of the variance of the predicted values to the variance of the observed values. While $R^2$ provides a useful measure of model fit, it should be interpreted cautiously, especially when comparing models with different numbers of predictors or when working with small sample sizes. In such cases, the adjusted $R^2$ or other information criteria like AIC may provide more reliable measures of model quality, and will be described following sections.

\subsection{Performing Regression in Practice}
\subsubsection*{Notation}
When performing any OLS regression of a variable $Y$ on the variables $X_1, ..., X_K$, we will denote the regression using the R-style formula:
\begin{definitionbox}{R-style Notation}
    The Linear Regression Model equation \ref{eq:linear} is denoted by:
    \begin{equation}
        \texttt{Y} \sim \texttt{X}_1 + ... + \texttt{X}_K
    \end{equation}
\end{definitionbox}

This notation omits the intercept, the coefficients, and the residuals, to clarify what are endogenous (\texttt{Y}) and exogenous ($\texttt{X}_k$) variables. 
\subsubsection*{Categorical Variables}
In some cases, a variable $X$ can be categorical instead of being continuous. In this case, if $X$ take $C$ discrete unordered values $D_1, ..., D_C$, we employ a \textit{\textbf{dummy coding}}: 
\begin{itemize}[noitemsep]
    \item Take a value, for example $D_1$, as a reference value
    \item Encode $X$ as a one-hot encoding vector in $\{0, 1\}^{C-1}$, with a 1 in the position $j$ corresponding to the observation of the value $D_{j}$.
    \item We get $C-1$ coefficients $\beta_2, ..., \beta_{C}$, corresponding to the \textbf{marginal effect of $X$ taking the value $D_{j}$ instead of the reference value $D_{1}$}, i.e. $$\beta_k = \mathds{1}(Y|X=D_k) - \mathds{1}(Y|X=D_1)$$
\end{itemize}
\subsubsection*{Product Variables}
In a regression of the form $\texttt{Y} \sim \texttt{X}_1 + \texttt{X}_2$, it is possible to model the case where the value of one variable can impact the marginal effect of the other one. For example, we might want to consider that the effect of training time on model performance depends on the model's complexity. This can be achieved by including an interaction term:
\begin{definitionbox}{Definition: Product variables}
To model interaction between variables, the equation 
\begin{equation}
   Y = \beta_0 + \beta_1X_1 + \beta_2X_2 + \beta_3(X_1 \cdot X_2) + \varepsilon
\end{equation}
is denoted equivalently by the formulas
\begin{equation}
\texttt{Y} \sim \texttt{X}_1 + \texttt{X}_2 + \texttt{X}_1:\texttt{X}_2 \Longleftrightarrow \texttt{Y} \sim \texttt{X}_1 \times \texttt{X}_2
\end{equation}
where $\texttt{X}_1:\texttt{X}_2$ represents the interaction between $\texttt{X}_1$ and $\texttt{X}_2$. 
\end{definitionbox}
To get sound interpretations, it is best to first center the exogenous variables, so $X_1 \cdot X_2$ is less correlated with $X_1$ and $X_2$.
In this case:
\begin{itemize}[noitemsep]
\item $\beta_1$ represents the effect of $X_1$ when $X_2$ is at its mean
\item $\beta_2$ represents the effect of $X_2$ when $X_1$ is at its mean
\item $\beta_3$ represents how the effect of $X_1$ changes for each unit increase in $X_2$ (and vice versa)
\end{itemize}
This allows us to model quite complicated dependencies between explanatory variables and the target variable.

Plain OLS regressions can be performed and interpreted as long as $Y$ is a continuous variable and that the CLRM hypothesis holds. The difficulty in a rigorous evaluation protocol does not come from complicated evaluation models but from a rigorous choice of explanatory variables, hypothesis verification, and careful interpretation of the results.
\section{Analysis of Variance (ANOVA)}
OLS regressions allow us to get marginal effects associated with each variable, which can be quite informative when the explanatory variables are continuous. For categorical variables, while dummy coding allows us to still get marginal effects from one group to another, it does not allow us to easily know what categorical variable is the most important, i.e. for which categorical variable the variance in the category is the greatest. For example, if we evaluate models by varying both model types and training types (both categorical variables), we need a way to say that one aspect has a greater influence on performance than the other.
\subsection{ANOVA as an Extension of OLS}
Analysis of Variance (ANOVA) is a powerful statistical technique used to analyze the differences among group means in a sample. While originally developed for experimental design, ANOVA has found wide applications in various fields, including machine learning evaluation.  ANOVA can be viewed as a special case of the Ordinary Least Squares (OLS) regression we discussed earlier. In fact, ANOVA and linear regression are two faces of the same coin, both falling under the General Linear Model framework.

Consider a model with two categorical variables $X_1$ and $X_2$:
\begin{equation}
Y = \beta_0 + \sum_{i} \beta_{1i} D_{1i} + \sum_j \beta_{2j} D_{2j} + \varepsilon
\end{equation}
Where $D_{1i}$ and $D_{2j}$ are dummy variables for $X_1$ and $X_2$, respectively. This OLS formulation is equivalent to the ANOVA model:
\begin{equation}
Y_{ij} = \mu + A_i + B_j + \varepsilon
\end{equation}
where $Y_{ij}$ is the response variable when $X_1 = D_{1i}$ and $X_2 = D_{2j}$,  $\mu$ is the overall mean effect, $A_i$ is the effect of $D_{1i}$, and $B_j$ is the effect of $D_{2j}$. We have $\mu + \alpha_1 + \beta_1 = \beta_0$ (for the reference categories), $A_i= \beta_{1j}$ for $i = 2, ..., J$

The ANOVA formulation can be advantageous since it naturally partitions the total variance into components associated with each factor and residual error. It allows for easier interpretation of main effects in the presence of multiple categorical variables, and provides a framework for analyzing complex experimental designs, including nested and crossed factors.

\subsection{Effect Sizes: Decomposing Model Variance}
While OLS focuses on estimating coefficients, ANOVA emphasizes decomposing the total variance in $Y$. This decomposition allows us to calculate effect sizes, particularly partial eta-squared $\eta^2_p$, which quantifies the proportion of variance explained by each factor.
\begin{definitionbox}{Definition: Partial $\eta^2$}
For a regression of the observed $y_i$ on the exogenous variables $x_{1i}, ..., x_{Ki}$, let $\hat{y}_i$ be the predicted outcome of the full model for the observation $i$, and $\hat{y}_{ki}$ the predicted outcome based only on $x_{ki}$. Then define the partial and error sum of squares by:
    \begin{align*}
        SS_{effect} &:= \sum_i (\hat{y}_{ki} - \hat{y}_i)^2\\
       SS_{error} &:= \sum_i (y_i - \hat{y}_i)^2
    \end{align*}
    Then the partial $\eta^2$, also denoted for $X_k$ is defined by:
    \begin{equation}
        \eta^2_p(X_k) = \frac{SS_{effect}}{SS_{effect}+SS_{error}}
    \end{equation}
\end{definitionbox}
\begin{resultbox}{Interpretation of $\eta^2_p$}
    $\eta^2_p$ represents the proportion of variance in $Y$ explained by a factor, after accounting for other factors in the model. It's directly related to the increase in $R^2$ when adding the factor to a model that already includes other factors.
\end{resultbox}

The ANOVA framework is particularly useful in machine learning evaluation when:
\begin{itemize}[noitemsep]
\item Comparing the effects of multiple categorical factors (e.g., model architecture, dataset type)
\item Quantifying the relative importance of different factors using $\eta^2_p$
\item Analyzing interaction effects between factors (in multi-way ANOVA)
\end{itemize}
While OLS provides coefficient estimates, ANOVA's focus on variance decomposition often provides a more intuitive understanding of each factor's importance in explaining variability in model performance.

\section{Binary Outcome Models: The Logit Model}
The general linear model is not adapted in the case where the endogenous variable $Y$ is binary. If $Y\in \{0,1\}$, then $\mathbb{E}[Y|X]\in[0,1]$. In the linear model, we assume a solution of the form $\mathbb{E}[Y|X]=X^T\beta$, but nothing guarantees that $X^T\beta$ is in $[0,1]$, which raises serious specification questions.

Additionally, the usual interpretation of $\beta_k$ as a marginal effect becomes less clear. If $Y$ is binary, interpreting a $\beta_k$ value of, for example, $0.6$, is not easy. This is because the marginal effect on a binary outcome cannot be directly interpreted in the same way as it would for a continuous outcome, since the effect on the probability is nonlinear.

\subsection{Definition of the Logit Model}
A solution is to restrict the predictions to a function $F$ that maps the linear predictor to the $[0,1]$ interval, ensuring that the predicted values are valid probabilities. One such function is the logistic function, commonly used in the Logit model.
\begin{definitionbox}{Definition: Generalized Linear Model (GLM)}
    If $F$ is a strictly increasing bijective function from $\mathbb{R}$ to $]0,1[$, it is a cumulative distribution function, a \textbf{Generalized Linear Model} is specified as:
    \begin{equation}
        \mathbb{E}[Y|X] = F(X\beta)
    \end{equation}
\end{definitionbox}
In particular, this allows us to model non-linearities. In the binary outcome case, we can consider that there exists a latent variable $Y^*$, such that $Y^* = X\beta + \varepsilon$, where $-\varepsilon$ has a c.d.f $F$, and $ Y= \mathds{1}(Y^* \geq 0)$. Then $$\mathbb{E}[Y|X] = \mathbb{P}(Y=1|X) = \mathbb{P}(Y=1|X) =  \mathbb{P}(-\varepsilon\leq X\beta|X) = F(X\beta)$$.

A common choice of function $F$ is the sigmoid function, which gives rise to the Logit model:
\begin{definitionbox}{Definition: The Logit Model}
    Let $\sigma$ be the sigmoid function, i.e.
    \begin{equation}
        \forall x\in \mathbb{R}, \sigma(x) = \frac{1}{1+e^{-x}}
    \end{equation}
    Then the logit model is defined by :
        \begin{equation}
        \label{eq:logit}
        \mathbb{E}[Y|X] = \sigma(X\beta)
    \end{equation}
    In R-style notation, it will be denoted:
    \begin{equation}
        \texttt{Y} \sim \sigma(\texttt{X}_1+...+\texttt{X}_K)
    \end{equation}
\end{definitionbox}
\subsection{Fitting the Model}
Sadly, there is no closed-form solution for the $\beta$ coefficients in the equation \ref{eq:logit} as in the OLS formulation. Thus, in practice, the model is fitted using the Maximum Likelyhood Estimation (MLE). The likelihood function for n independent observations is:
\begin{equation}
L(\beta) = \prod_{i=1}^n [\sigma(x_i\beta)]^{y_i} [1-\sigma(x_i\beta)]^{1-y_i}
\end{equation}
We maximize the log-likelihood:
\begin{equation}
\ell(\beta) = \sum_{i=1}^n y_i \log(\sigma(x_i\beta)) + (1-y_i) \log(1-\sigma(x_i\beta))
\end{equation}
This is typically done using numerical optimization methods like Newton-Raphson or Fisher scoring.

To obtain confidence intervals for $\beta$, we use the fact that the MLE estimator $\hat{\beta}$ is asymptotically normal. The variance-covariance matrix of  $\hat{\beta}$ can be estimated using the inverse of the observed Fisher information matrix:
\begin{equation}
\widehat{Var}(\hat{\beta}) = I(\hat{\beta})^{-1} = \left(-\frac{\partial^2 \ell(\beta)}{\partial \beta \partial \beta^T}\bigg|_{\beta=\hat{\beta}}\right)^{-1}
\end{equation}
This allows us to construct confidence intervals:
\begin{resultbox}{Confidence Intervals for the Logit Model}
    For a given coefficient $\beta_k$, let $z_{1-\alpha/2}$ be the $1-\alpha/2$-th quantile for the Gaussian distribution. Then, with probability $1-\alpha$:
\begin{equation}
\beta_k \in \left[\widehat{\beta}_k \pm z_{1-\alpha/2} \sqrt{\frac{\widehat{Var}(\hat{\beta})_{k,k}}{n}}\right]
\end{equation}
where $\widehat{Var}(\hat{\beta})_{k,k}$ is the $k$-th diagonal element of $\widehat{Var}(\hat{\beta})$.
\end{resultbox}
\begin{proof}
    The MLE estimator $\hat{\beta}$ is asymptotically normal with distribution $\mathcal{N}(\beta, \frac{1}{n}I(\beta)^{-1})$ where $I(\beta)$ is the Fisher information matrix evaluated at $\beta$. Evaluating the Fisher information matrix at the empirical counterpart $\hat{\beta}$ yields the desired confidence interval.
\end{proof}
\subsection{Interpreting the Model}
\subsubsection*{Marginal Effects}
Reversing the equation \ref{eq:logit}, we can the equivalent model:
\begin{equation}
    \ln \frac{\mathbb{P}(Y=1|X)}{\mathbb{P}(Y=0|X)} = X\beta
\end{equation}
Therefore, the logit regression can be viewed as a linear regression in the log-odds ratio scale. This makes it quite difficult to directly interpret the values $\beta_k$ as concrete effects, since we would want to have an interpretation on the probability $\mathbb{P}(Y|X)$. For the logit model, we get :
\begin{equation}
\frac{\partial \mathbb{P}(Y=1|X)}{\partial X_k} = \sigma(X\beta)(1-\sigma(X\beta))\beta_k
\end{equation}
Contrary to the OLS, the marginal effect of $X_k$ on the probability of outcome $Y$ depends on all the other variables $X_{-k}$. A common quantity to consider is then the Average Marginal Effect:
\begin{definitionbox}{Definition: Average Marginal Effect}
    In the logit model, the \textbf{Average Marginal Effect (AME)} of a variable $X_k$ on the probability of outcome $Y$ is:
    \begin{equation}
        AME(X_k) = \mathbb{E}\left[\frac{\partial \mathbb{P}(Y=1|X)}{\partial X_k}\right]
    \end{equation}
    It represents the effect of a unit change in $X_k$, on average, on the probability of outcome $Y$.
    Its empirical counterpart is estimated by:
    \begin{equation}
        \widehat{AME}(X_k) =\frac{1}{n}\sum_{i=1}^n\frac{\partial \widehat{\mathbb{P}}(Y=1|X)}{\partial X_k}\bigg|_{X=x_i} = \frac{1}{n}\sum_{i=1}^n \sigma(x_i\hat{\beta})(1-\sigma(x_i\hat{\beta}))\hat{\beta}_k
    \end{equation}
\end{definitionbox}
The confidence intervals for $AME(X_k)$ are computed using the delta method. If $g_k(\beta)$ is the function that computes the marginal effector $X_k$, then:
\begin{equation}
\sqrt{n}(g_k(\hat{\beta}) - g_k(\beta)) \overset{\mathcal{L}}{\longrightarrow} \mathcal{N}(0, \nabla g_k(\beta)^T I(\beta)^{-1} \nabla g_k(\beta))
\end{equation}
The variance of the marginal effect estimate is then approximated as:
\begin{equation}
\widehat{Var}(g_k(\hat{\beta})) = \nabla g_k(\hat{\beta})^T \hat{I}(\hat{\beta})^{-1} \nabla g_k(\hat{\beta})
\end{equation}
This allows us to construct confidence intervals for the marginal effects:
\begin{resultbox}{Confidence Intervals of the AME}
     Let $z_{1-\alpha/2}$ be the $1-\alpha/2$-th quantile for the Gaussian distribution. Then, with probability $1-\alpha$:
     \begin{equation}
         AME(X_k) \in \left[\widehat{AME}(X_k) \pm z_{1-\alpha/2} \sqrt{\frac{\widehat{Var}(g_k(\hat{\beta}))}{n}}\right]
     \end{equation}
\end{resultbox}
\subsubsection*{Goodness of Fit}
In logistic regression, the standard $R^2$ used in linear regression is not applicable due to the non-linear nature of the model. McFadden's $R^2$, also known as the likelihood ratio index, is one of several pseudo-$R^2$ measures developed for logistic regression and other models estimated by maximum likelihood.
\begin{definitionbox}{Definition: McFadden's $R^2$}
For a logit model, McFadden's $R^2$ is defined as:
\begin{equation}
R^2_{\text{McFadden}} = 1 - \frac{\ln(L_{\text{full}})}{\ln(L_{\text{null}})}
\end{equation}
Where:
\begin{itemize}
\item $L_{\text{full}}$ is the likelihood of the full model
\item $L_{\text{null}}$ is the likelihood of the null model (intercept-only model)
\end{itemize}
\end{definitionbox}
Unlike the $R^2$ in linear regression, McFadden's $R^2$ does not have an upper bound of 1, and values between 0.2 and 0.4 are considered to represent an excellent fit.
However, like other pseudo-$R^2$ measures, McFadden's $R^2$ should be used in conjunction with other diagnostic tools and substantive interpretation of the model coefficients. It's particularly useful when the goal is to understand the factors influencing binary outcomes in machine learning models, rather than purely predictive tasks.

\section{Model Selection and Validation}
In the context of evaluation and explainability using regressions, selecting the most appropriate statistical model is required for drawing valid conclusions. This section discusses key tools for model selection and validation, focusing on information criteria and residual analysis.
\subsection*{Information Criteria}
While the $R^2$ of a regression gives us an idea of the model explainability power, it suffers from problems, particularly overfitting. To select the best model, the \textbf{Akaike Information Criterion (AIC)} is a widely used tool for model selection, balancing model fit against complexity to avoid overfitting.
\begin{definitionbox}{Definition: Akaike Information Criterion (AIC)}
For a model with $K$ parameters and maximum likelihood $L$, the AIC is defined as:
\begin{equation}
AIC = 2K - 2\ln(L)
\end{equation}
Where:
\begin{itemize}[noitemsep]
\item $K$ is the number of estimated parameters in the model
\item $L$ is the maximum value of the likelihood function for the model
\end{itemize}
\end{definitionbox}
The AIC is founded on information theory and provides a relative estimate of the information lost when a given model is used to represent the process that generates the data. When comparing models, the one with the lower AIC is generally preferred. The AIC has some interesting properties:
\begin{itemize}[noitemsep]
    \item \textbf{Trade-off}: AIC rewards goodness of fit (as assessed by the likelihood function) but includes a penalty that increases with the number of estimated parameters. This penalty discourages overfitting.
    \item \textbf{Relative measure}: AIC values are only meaningful when compared between models. The absolute value of AIC for a single model is not interpretable on its own.
    \item \textbf{Model comparison}: When comparing models, a difference in AIC of 2 is often considered the threshold for meaningful difference.
\end{itemize}

\subsection*{Residual Analysis}
 To ensure the validity of our statistical inferences, we need to verify the regression assumptions. Since the theoretical model cannot be accessed, the key OLS assumptions are checked through various residual plots and statistical tests, called \textbf{regression diagnostics}. Recall that the residuals are defined by:
 \begin{equation}
     \epsilon_i = y_i - \hat{y}_i
 \end{equation}
The key assumptions are verified using different methods:
\begin{itemize}[noitemsep]
    \item \textbf{Linearity}: Plot the $\epsilon_i$ vs. the fitted values $\hat{y}_i$, and verify it is centered around 0. The idea is that to check if $y_i$ is linear in $x_i$ and that all the linear effects have been accounted for, we can check if $\mathbb{E}[\epsilon_i|x_i] = 0$. In particular, any pattern in this plot, such as a U-shaped curve, indicates missing non-linear effects
    \item \textbf{Normality} of residuals: Plot the Q-Q plot of the standardized $\epsilon_i$ against the Gaussian distribution. Tests like the  Kolmogorov-Smirnov (KS) can be used to check if the residuals are indeed Gaussian, but should be used with caution: for a sample large enough, a small deviation from the Gaussian distribution, even if not too problematic,  can be detected by the KS test. A graphical inspection can be more insightful in these cases.
    \item \textbf{No multicollinearity} among predictors: Compute each predictor's Variance Inflation Factors (VIF), which indicates how much a predictor is linear in the others. A VIF value greater than 5 or 10 indicates problematic multicollinearity.
    \item \textbf{Homoscedasticity} (constant variance of residuals): plot $\epsilon_i$ vs $\hat{y}_i$ or $\sqrt{\Tilde{\epsilon}_i}$ vs $\hat{y}_i$, where $\Tilde{\epsilon}_i$ are the re-scaled residuals. The distribution should be the same for all values of $\hat{y}_i$. If the homoscedasticity assumption does not hold, the value of the $\beta_k$ coefficients is still valid, but the confidence intervals become imprecise. This can be solved by using more robust and more conservative confidence intervals. 
    \item \textbf{No influential outliers}: The influence of an observation $i$ on the model can be measured by its \textbf{leverage} $\hat{\boldsymbol{H}}_{ii}$ i.e. the $i$-th diagonal element of the hat matrix. Plotting the leverage vs. the residual value highlights aœny outlier point. Additional measures, like Cook's distance, can be used to quantify the effect of each observation.
\end{itemize}
\subsubsection*{Residuals Analysis for the Logit Model}
While the overall principles of model diagnostics remain important for logit regressions, the methods used for linear regression cannot be directly applied due to the non-linear nature of the logit model. Indeed, in linear regression, residuals are expected to be normally distributed with constant variance. However, in logistic regression:
\begin{itemize}[noitemsep]
    \item \textbf{Non-normality}: The residuals in logistic regression are not normally distributed. Instead, they follow a binomial distribution.
    \item \textbf{Heteroscedasticity}: The variance of residuals is not constant but depends on the predicted probability.
    \item \textbf{Bounded nature}: Residuals in logistic regression are bounded, unlike in linear regression where they can take any value.
\end{itemize}
These characteristics make traditional residual plots (like residuals vs. fitted values) less informative and potentially misleading for logistic regression.

Several methods have been proposed to find counterparts to the linear diagnostics for the logit model, like deviance residuals or Pearson residuals \citep{pierce1982residuals, mccullagh1989generalized}. However, in practice, they often fall short of providing a clear understanding are the practical problems in a logit model.

DHARMa (Diagnostics for HierArchical Regression Models) \citep{DHARMa} is a method designed to create readily interpretable residuals for generalized linear mixed models (GLMMs), including logistic regression. The key ideas of DHARMA are:
\begin{itemize}[noitemsep]
    \item \textbf{Simulation-based approach}: DHARMA simulates new data from the fitted model multiple times.
    \item \textbf{Quantile residuals}: It calculates the quantile of the observed data within the simulated data distribution.
    \item \textbf{Uniformity expectation}: If the model is correctly specified, these quantile residuals should follow a uniform distribution.
\end{itemize}
The generated DHARMa residuals can be used to verify all the classical diagnostics plots presented in the previous subsection. This allows us to overcome the limitations of traditional residual analysis in logistic regression and gain insights into model fit and potential issues. This approach provides a robust method for assessing the validity of our logistic regression models used in binary classification tasks or when analyzing binary outcomes in ML experiments.

\section{Examples of Applications}
\subsection{Analysis of Pre-training Strategies for Class-Incremental Learning}
This study by \citep{petit2024analysis} presents a comprehensive analysis of initial training strategies for Exemplar-Free Class-Incremental Learning (EFCIL). The authors investigate how different pre-training methods, neural architectures, and EFCIL algorithms impact performance across various datasets. They employ a rigorous statistical framework to quantify the influence of these factors on incremental learning performance.
\subsubsection{Experimental Setup}
Several training strategies are compared:
\begin{itemize}
    \item Supervised learning on the initial batch of target data
    \item Self-supervised learning (SSL) methods like MoCov3, BYOL, and DINOv2
    \item Pre-training on external datasets (e.g., ImageNet)
    \item Fine-tuning pre-trained models on the initial batch
\end{itemize}

This comparisons are performed on two common architectures, ResNet50 \citep{he2016_resnet} and Vision Transformers \citep{dosovitskiy2020image}.

Three EFCIL algorithms are tested:
\begin{itemize}
    \item BSIL \citep{jodelet2021balanced} : a fine-tuning based method
\item DSLDA \citep{hayes2020_deepslda}: a transfer learning based methods that freeze the feature extractor
\item FeTrIL \citep{fetril_Petit_2023_WACV}: another more recent transfer learning approach
\end{itemize}

The experiments cover 16 target datasets and two EFCIL scenarios, resulting in 1,248 total experiments.

\subsubsection{Statistical Analysis}
The authors use Ordinary Least Squares (OLS) regression and Analysis of Variance (ANOVA) to model the causal effects of different factors on EFCIL performance. They consider metrics such as average incremental accuracy and forgetting.

\begin{resultbox}{Findings of a Rigorous Analysis of EFCIL}
    The findings from the statistical analysis include:
\begin{itemize}
    \item The initial training strategy is the dominant factor influencing average incremental accuracy.
\item The choice of EFCIL algorithm is more important in preventing forgetting.
\item Pre-training with external data generally improves accuracy.
\item Self-supervised learning in the initial step boosts incremental learning, especially when the pre-trained model is fine-tuned on initial classes.
\item Transfer learning-based EFCIL algorithms outperform fine-tuning-based methods in most cases.
\end{itemize}
\end{resultbox}

This allows the authors to draw concrete recommendations for the evolution of the EFCIL field. Overall, the effectiveness of pre-training depends on the similarity between the source and target datasets.
Fine-tuning pre-trained models is generally beneficial for CNNs but can be detrimental for transformers in EFCIL, whereas Self-supervised learning is particularly useful when the amount of initial data is limited.
The choice between CNNs and transformers doesn't significantly affect performance, suggesting both architectures should be explored in future EFCIL work.

\subsection{Toward Fairer Face Recognition Datasets}
This study by \citet{fourniermongieux2024fairerfacerecognitiondatasets} addresses the issue of fairness in face recognition and verification tasks. The authors propose a novel method to balance demographic attributes in generated training datasets and introduce a comprehensive evaluation framework that considers both accuracy and fairness. This work aims to mitigate biases in face recognition systems while maintaining high performance.

\subsubsection{Experimental Setup}
The authors compare several face recognition training datasets:
\begin{itemize}
    \item Real dataset: CASIA \citep{yi2014casia}
\item Synthetic datasets: SynFace\citep{qiu2021synface}, DigiFace\citep{bae2023digiface}, DCFace\citep{kim2023dcface}
\item Novel Proposed balanced synthetic datasets:  DCFace-eg, DCFace-All
\end{itemize}
Each dataset contains 10,000 identities with 50 images per identity and models are trained using a ResNet50 backbone with the AdaFace method. Evaluation is performed on seven face verification datasets, including challenging ones like LFW-C and FAVCI2D, to assess both accuracy and fairness.

The study focuses on four main demographic attributes:
\begin{itemize}
    \item Gender: Male, Female
\item  Age: Young, Adult, Senior
\item Ethnicity: Asian, Black, Indian, White
\item  Pose: 3D angles (pitch, yaw, roll)
\end{itemize}

This setup allows the authors to compare biases across different dataset generation methods (real vs. synthetic) and assess the impact of their proposed balancing technique on fairness in face recognition systems.

\subsubsection*{Statistical Analysis}
The authors employ two main statistical techniques to analyze fairness in face recognition systems:
\begin{enumerate}
    \item Logistic Regression is used to model the probability of correct identification (True Positive Rate and False Positive Rate). This allows the estimation of marginal effects of demographic attributes on model performance and helps identify statistically significant differences in performance across demographic groups.
\item ANOVA is used to decompose the variance in the embedding space and quantifies the contribution of different factors (ethnicity, age, gender, pose) to overall variance. This Helps identify which factors have the strongest influence on the model's behavior in the latent space.
\end{enumerate}
These techniques allow for the isolation of individual attribute effects while controlling for others, quantification of bias with statistical significance, and an analysis of both the embedding space and classification outcomes.

\begin{resultbox}{Findings of a Rigorous Measure of Biases in Face Verification}
    Embedding Space Analysis:
\begin{itemize}
    \item Pose differences strongly influence performance for positive pairs
    \item Ethnicity and age have a greater impact on performance for negative pairs
    \item Unbalanced datasets lead to an unequal population of the embedding space
\end{itemize}

\noindent True Positive Rate (TPR) and False Positive Rate (FPR) Analysis:
\begin{itemize}
    \item All models exhibit some unfairness regarding ethnicity in TPR
    \item Balanced datasets show improved fairness between ethnic groups in FPR
    \item Controlling multiple attributes simultaneously is important for reducing overall bias
\end{itemize}

\end{resultbox}

This analysis allows the authors to show that the proposed balancing method significantly reduces demographic unfairness.
A performance gap persists between real and synthetic datasets, but to is narrowing. According to the authors, the DCFace-All dataset achieves the best balance between accuracy and fairness

This rigorous statistical approach provides a more nuanced understanding of biases in face recognition systems, revealing subtle effects that might be missed by simpler evaluation methods.

\section{Conclusion}
We have presented a comprehensive overview of statistical tools that can significantly enhance the analysis and evaluation of machine learning models. By adapting econometric methods to the context of machine learning, we have demonstrated how researchers can move beyond simple performance metrics to gain deeper insights into model behavior and performance.

The ordinary least squares (OLS) regression provides a fundamental framework for understanding the relationships between various factors and model performance. Its extension to Analysis of Variance (ANOVA) offers a powerful tool for assessing the relative importance of different categorical variables in explaining model behavior. For scenarios involving binary outcomes, often encountered in classification tasks, the logistic regression model offers a robust approach to analysis.

These statistical techniques offer several key advantages in the context of machine learning evaluation. Under appropriate assumptions, these methods allow us to draw causal conclusions about the factors influencing model performance, going beyond mere correlation. Moreover, the confidence intervals and hypothesis tests associated with these methods provide a rigorous framework for assessing the reliability of our findings.

However, it's important to note that these methods are not without limitations. The assumptions underlying these techniques (such as linearity, homoscedasticity, and normality of residuals) may not always hold in the context of complex machine learning models.
Despite these challenges, the rigorous application of these statistical tools can significantly enhance our understanding of machine learning models. They provide a complementary perspective to traditional machine learning evaluation metrics, offering insights into not just how well a model performs, but why it performs as it does.

\setlength{\baselineskip}{0pt} 
\newpage
\bibliography{references}

\end{document}